\documentclass[acmsmall,nonacm]{acmart}
\usepackage[utf8]{inputenc}
\usepackage{tabularx}
\usepackage{graphicx}
\usepackage{booktabs}
\usepackage{multirow}
\usepackage{amsmath}
\usepackage{url} 
\usepackage{tikz} 
\usepackage{varwidth} 
\usepackage{hyperref} 
\usepackage{color}
\usepackage{todonotes} 
\usepackage{lscape}
\usepackage{longtable}
\usepackage{adjustbox}
\usepackage{pifont}

\usepackage{etoolbox}
\makeatletter
\patchcmd{\maketitle}{\@copyrightspace}{}{}{}
\makeatother

\newcommand{\bigyes}{\ding{52}\ding{52}}
\newcommand{\yes}{\ding{51}}
\newcommand{\no}{\ding{55}}

\tikzset{
  basic/.style  = {draw, text width=2cm, drop shadow, font=\sffamily, rectangle},
  root/.style   = {basic, rounded corners=2pt, thin, align=center,
                   fill=green!30},
  level 2/.style = {basic, rounded corners=6pt, thin,align=center, fill=green!60,
                   text width=8em},
  level 3/.style = {basic, thin, align=left, fill=pink!60, text width=6.5em}
}

\title{Technologies for Trustworthy Machine Learning: A Survey in a Socio-Technical Context}

\author{Ehsan Toreini}
\email{ehsan.toreini@durham.ac.uk}
\affiliation{
  \institution{Department of Computer Science, Durham University}
  \city{Durham}
  \state{United Kingdom}
}
\author{Mhairi Aitken}
\email{maitken@turing.ac.uk}
\affiliation{
  \institution{Alan Turing Institute}
  \city{London}
  \state{United Kingdom}
}
\author{Kovila P. L. Coopamootoo}
\email{kovila.coopamootoo@newcastle.ac.uk}
\affiliation{
  \institution{School of Computing, Newcastle University}
  \city{Newcastle upon Tyne}
  \state{United Kingdom}
}
\author{Karen Elliott}
\email{karen.elliott@newcastle.ac.uk}
\affiliation{
  \institution{Newcastle University Business School}
  \city{Newcastle upon Tyne}
  \state{United Kingdom}
}
\author{Vladimiro Gonzalez Zelaya}
\email{c.v.gonzalez-zelaya2@newcastle.ac.uk}
\affiliation{
  \institution{School of Computing, Newcastle University}
  \city{Newcastle upon Tyne}
  \state{United Kingdom}
}
\author{Paolo Missier}
\email{paolo.missier@newcastle.ac.uk}
\affiliation{
  \institution{School of Computing, Newcastle University}
  \city{Newcastle upon Tyne}
  \state{United Kingdom}
}
\author{Magdalene Ng}
\email{magdalene.ng@newcastle.ac.uk}
\affiliation{
  \institution{School of Computing, Newcastle University}
  \city{Newcastle upon Tyne}
  \state{United Kingdom}
}
\author{Aad van Moorsel}
\email{aad.vanmoorsel@newcastle.ac.uk}
\affiliation{
  \institution{School of Computing, Newcastle University}
  \city{Newcastle upon Tyne}
  \state{United Kingdom}
}

\renewcommand{\shortauthors}{Toreini, Aitken, Coopamootoo, Elliott, Zelaya, Missier, Ng and Van Moorsel}

\begin{document}

\begin{abstract}
Concerns about the societal impact of AI-based services and systems has encouraged governments and other organisations around the world to propose AI policy frameworks to address fairness, accountability, transparency and related topics. To achieve the objectives of these frameworks, the data and software engineers who build machine-learning systems require knowledge about a variety of relevant supporting tools and techniques. In this paper we provide an overview of technologies that support building trustworthy machine learning systems, i.e., systems whose properties justify that people place trust in them.  We argue that four categories of system properties are instrumental in achieving the policy objectives, namely fairness, explainability, auditability and safety \& security (FEAS). We discuss how these properties need to be considered across all stages of the machine learning life cycle, from data collection through run-time model inference. As a consequence, we survey in this paper the main technologies with respect to all four of the FEAS properties, for data-centric as well as model-centric stages of the machine learning system life cycle.  We conclude with an identification of open research problems, with a particular focus on the connection between trustworthy machine learning technologies and their implications for individuals and society.     
\end{abstract}




\maketitle

\section{Introduction}
\label{s:intro}
Recent years have seen a proliferation of research in technologies for trustworthy machine learning as well as in policy frameworks that introduce requirements around fairness, human rights, privacy and other properties of AI-based services~\cite{aitken2020establishing}.  When building AI-based services, systems and products, the challenge data, software and system engineers face is to design systems that deliver the societal quality objectives articulated in the policy frameworks.  In this paper, we consider this challenge in building AI-based systems specifically with respect to machine learning technologies. We do this through a survey that considers technologies for trustworthy machine learning in relation to emerging policy frameworks. 

To set the stage, we review and discuss a set of prominent policy frameworks for AI. Such policy frameworks are emerging in many nations, jurisdictions and organisations, to respond to societal concerns about the impact of AI on the quality of life of individuals and on our society, for instance in terms of fairness and justice.  There is a noticeable and understandable gap between the subject of the discourse in AI policy frameworks and that of technical approaches in the machine learning literature. Typical concerns in the policy frameworks are expressed in terms such as ethics, justice, fairness, accountability and transparency. These do not map one-to-one to properties of trustworthy systems.  (Here, the term trustworthy indicates the existence of system properties that justify parties to place trust in the system \cite{Avizienis04}). 

We argue that the policy framework objectives can be implemented through four key properties of trustworthy machine learning systems: Fairness, Explainability, Auditability and Safety \& Security (FEAS).  The survey provides an encompassing view of machine learning technologies available with respect to each of the FEAS properties. Of course, each FEAS property deserves its own literature review, the objective in this paper is to highlight the main characteristics and types of approaches for each property to provide a sounds foundation for machine learning systems engineering research and development. To further maintain a reasonable scope, within the safety \& security category the emphasis is on security and privacy, more so than on safety, reliability and similar concerns.  

To build trustworthy machine learning systems, one needs to incorporate concerns about trustworthiness in all stages of the design, from data collection strategies to algorithm selection and visualisation of results. Similarly, these should be considered in all life cycle stages, including design, development and deployment. We use the chain of trust, introduced in \cite{toreini2020relationship}, to illustrate and discuss the dependencies between stages of the system design and life cycle. In particular, we organise our survey in two classes, Data-Centric Trustworthiness (data collection, pre--processing and feature engineering) and Model-Centric Trustworthiness (model training, testing and inference).  

We conclude the paper with a discussion of open problems in building trustworthy machine learning systems.  The open problems concentrates on the main research required to establish an implementation pathway for machine-learning based systems to fulfil qualities emerging from the AI policy framework. The future research considers both technology and socio-technical research questions.

The paper is organised as follows. In Section \ref{s:FEAS}, we discuss the relationship between emerging policy framework for AI and the four technology FEAS properties. Section \ref{s:chain} the introduces the various stages in the system life cycle at which trustworthiness is to be considered, based on the chain of trust. Section \ref{s:challenges} sets the stage for the discussion of FEAS technologies by identifying challenges in trustworthy machine learning and introducing definitions and ways to assess the properties.  The core of the survey is in the next two section, first surveying data-centric technologies in Section \ref{s:data} and then model-centric technologies in Section \ref{s:model}. Section \ref{s:open} discusses open research problems to bridge the gap between technologies and the desired societal quality objectives, and Section \ref{s:conclusion} concludes the paper.  

{\tiny
	

\begin{table*}[]
\caption{Trustworthy technology classes related to FAT* frameworks. \no = \emph{no mention}, \yes = \emph{mentioned}, \bigyes = \emph{emphasised}}
\begin{tabular}{@{}l|clcccccc@{}}
\toprule
	                                                                       Framework                                 & Year & Document Owner                                                                                                                         & Entities                                                  & Country        & Fairness & Explainability & Safety & Auditability \\ \midrule
	Top 10 principles of ethical AI                                                                         & 2017 & UNI Global Union                                                                                                                       & Ind                                                       & Switzerland    & \yes      & \yes            & \yes      & \yes          \\ \midrule
	Toronto Declaration                                                                                     & 2018 & Amnesty International                                                                                                                  & Gov, Ind                                                  & Canada         & \bigyes   & \yes            & \no       & \yes          \\ \midrule
	\begin{tabular}[c]{@{}l@{}}Future of work and Education\\ For the Digital Age\end{tabular}              & 2018 & T20: Think 20                                                                                                                          & Gov                                                       & Argentina      & \bigyes   & \yes            & \yes      & \yes          \\ \midrule
	Universal Guidelines for AI                                                                              & 2018 & The public voice coalition                                                                                                             & Ind                                                       & Belgium        & \bigyes   & \yes            & \yes      & \yes          \\ \midrule
	Human Rights in the Age of AI                                                                           & 2018 & Access Now                                                                                                                             & Gov, Ind                                                  & United States  & \bigyes   & \yes            & \bigyes   & \yes          \\ \midrule
	Preparing for the Future of AI                                                                          & 2016 & \begin{tabular}[c]{@{}l@{}}US national Science,\\ and Technology Council\end{tabular}                                                  & \begin{tabular}[c]{@{}l@{}}Gov, Ind, \\ Acad\end{tabular} & United States  & \yes      & \yes            & \bigyes   & \yes          \\ \midrule
	Draft AI R\&D Guidelines                                                                                 & 2017 & Japan Government                                                                                                                       & Gov                                                       & Japan          & \no       & \yes            & \bigyes   & \yes          \\ \midrule
	White Paper on AI Standardization                                                                     & 2018 & \begin{tabular}[c]{@{}l@{}}Standards Administration \\ of China\end{tabular}                                                                                                     & Gov, Ind                                                  & China          & \yes      & \no             & \bigyes   & \bigyes       \\ \midrule
	\begin{tabular}[c]{@{}l@{}}Statements on AI, Robotics and \\ `Autonomous' Systems\end{tabular}          & 2018 & \begin{tabular}[c]{@{}l@{}}European Group \\ on Ethics in Science \\ and New Technologies\end{tabular}                                    & \begin{tabular}[c]{@{}l@{}}Gov, Ind,\\ Acad\end{tabular}  & Belgium        & \yes      & \yes            & \yes      & \bigyes       \\ \midrule
	\begin{tabular}[c]{@{}l@{}}For a Meaningful Artificial\\  Intelligence\end{tabular}                     & 2018 & \begin{tabular}[c]{@{}l@{}}Mission assigned by \\ the French Prime Minister\end{tabular}                                               & Gov, Ind                                                  & France         & \yes      & \yes            & \no       & \bigyes       \\ \midrule
	AI at the Service of Citizens                                                                           & 2018 & Agency for Digital Italy                                                                                                               & Gov, Ind                                                  & Italy          & \yes      & \yes            & \yes      & \yes          \\ \midrule
	AI for Europe                                                                                           & 2018 & European Commission                                                                                                                    & Gov, Ind                                                  & Belgium        & \yes      & \yes            & \yes      & \yes          \\ \midrule
	AI in the UK                                                                                            & 2018 & UK House of Lords                                                                                                                      & Gov, Ind                                                  & United Kingdom & \bigyes   & \yes            & \yes      & \yes          \\ \midrule
	AI in Mexico                                                                                            & 2018 & \begin{tabular}[c]{@{}l@{}}British Embassy \\ in Mexico City\end{tabular}                                                                                                         & Gov                                                       & Mexico         & \yes      & \no             & \yes      & \yes          \\ \midrule
	Artificial Intelligence Strategy                                                                        & 2018 & \begin{tabular}[c]{@{}l@{}}German Federal Ministries \\ of Education, Economic \\ Affairs, and Labour \\ and Social Affairs\end{tabular} & Gov, Ind                                                  & Germany        & \yes      & \yes            & \yes      & \bigyes       \\ \midrule
	\begin{tabular}[c]{@{}l@{}}Draft Ethics Guidelines for\\  Trustworthy AI\end{tabular}                    & 2018 & \begin{tabular}[c]{@{}l@{}}European High Level Expert \\ Group on AI\end{tabular}                                                      & Gov, Ind, Civ                                             &                & \bigyes   & \yes            & \bigyes   & \yes          \\ \midrule
	AI Principles and Ethics                                                                                 & 2019 & Smart Dubai                                                                                                                            & Ind                                                       & UAE            & \bigyes   & \yes            & \bigyes   & \yes          \\ \midrule
	\begin{tabular}[c]{@{}l@{}}Principles to Promote FEAT AI in \\ the Financial Sector\end{tabular}        & 2019 & \begin{tabular}[c]{@{}l@{}}Monetary Authority \\ of Singapore  \end{tabular}                                                                                                      & Gov, Ind                                                  & Singapore      & \yes      & \yes            & \no       & \yes          \\ \midrule
	Tenets                                                                                                  & 2016 & Partnership on AI                                                                                                                      & Gov, Ind, Acad                                            & United States  & \yes      & \yes            & \yes      & \yes          \\ \midrule
	Asilomar AI Principles                                                                                  & 2017 & Future of Life Institute                                                                                                               & Ind                                                       & United States  & \yes      & \yes            & \yes      & \yes          \\ \midrule
	The GNI Principles                                                                                      & 2017 & Global Network Initiative                                                                                                              & Gov, Ind                                                  & United States  & \no       & \yes            & \yes      & \yes          \\ \midrule
	Montreal Declaration                                                                                    & 2018 & University of Montreal                                                                                                                 & Gov, Ind, Civ                                             & Canada         & \bigyes   & \bigyes         & \bigyes   & \yes          \\ \midrule
	Ethically Aligned Design                                                                                & 2019 & IEEE                                                                                                                                   & Ind                                                       & United States  & \yes      & \yes            & \yes      & \bigyes       \\ \midrule
	Seeking Ground Rules for AI                                                                             & 2019 & New York Times                                                                                                                         & Ind, GeP                                                  & United States  & \yes      & \yes            & \yes      & \yes          \\ \midrule
	\begin{tabular}[c]{@{}l@{}}European Ethical Charter on \\ the Use of AI in Judicial Systems\end{tabular} & 2018 & Council of Europe: CEPEJ                                                                                                               & Gov                                                       & France         & \yes      & \yes            & \yes      & \yes          \\ \midrule
	AI Policy Principles                                                                                    & 2017 & ITI                                                                                                                                    & Gov, Ind                                                  & United States  & \yes      & \yes            & \bigyes   & \yes          \\ \midrule
	The Ethics of Code                                                                                      & 2017 & Sage                                                                                                                                   & Ind                                                       & United States  & \yes      & \no             & \no       & \yes          \\ \midrule
	Microsoft AI Principles                                                                                 & 2018 & Microsoft                                                                                                                              & Ind                                                       & United States  & \yes      & \yes            & \bigyes   & \yes          \\ \midrule
	AI at Google: Our Principles                                                                            & 2018 & Google                                                                                                                                 & Ind                                                       & United States  & \yes      & \yes            & \bigyes   & \yes          \\ \midrule
	AI Principles of Telef\textbackslash{}'onica                                                            & 2018 & Telef\textbackslash{}'onica                                                                                                            & Ind                                                       & Spain          & \yes      & \yes            & \yes      & \no           \\ \midrule
	\begin{tabular}[c]{@{}l@{}}Guiding Principles on Trusted \\ AI Ethics\end{tabular}                      & 2019 & Telia Company                                                                                                                          & Ind                                                       & Sweden         & \yes      & \yes            & \bigyes   & \yes          \\ \midrule
	\begin{tabular}[c]{@{}l@{}}Declaration of the Ethical \\ Principles for AI\end{tabular}                 & 2019 & IA Latam                                                                                                                               & Ind                                                       & Chile          & \yes      & \yes            & \bigyes   & \yes          \\ \bottomrule

\end{tabular}
\label{tbl:frameworks}
\end{table*}
}

\section{Relating AI Policy Frameworks to Technology Properties}
\label{s:FEAS}
The widespread integration of AI in services, products and systems has raised various concern in society about issues such as ethics, employment, legal implications and human rights.   To be able to address these concerns a variety of AI policy frameworks has been proposed by governments and organisations world-wide articulated which qualities AI-based systems should exhibit with respect to the identified societal concerns.  In this section we discuss the requirements these frameworks pose on AI-based services (Section \ref{s:policy}) and argue that four technology properties (Fairness, Explainability, Auditability, Safety/Security) should be central to the design of machine-learning systems to fulfil these requirements (Section \ref{ss:FEAS}).  

\subsection{AI Policy Frameworks}
\label{s:policy}
Table~\ref{tbl:frameworks} provides a set of policy frameworks that have been proposed up to and including 2019, either exclusively considering AI or for technologies in general, including AI. These frameworks have been introduced by a variety of stakeholders, including technology companies, government, professional and standardisation bodies and academic researchers. The AI elements of these policy frameworks present varying sets of qualities that AI-based systems should exhibit to handle concerns about discrimination, accountability, human rights, etc. By their nature, these documents discuss high--level objectives for the concerned systems as well as the involved science and technology, but do not go into specific technical implementation. 

Recent literature provides very interesting reviews of AI policy frameworks. Whittlestone et al.~\cite{whittlestone2019role} provide a critical analysis of frameworks for ethical AI. They point out the inconsistency across frameworks, for instance, frameworks may confuse the interpretation of qualities, may present conflicting definitions and qualities may be different across different documents. Particularly relevant, also for the underlying technologies, is that the frameworks often fail to realise dependencies between policy objectives (e.g. addressing discrimination may lead to unexpected privacy leakages).  

Fjeld et al.~\cite{ethicalGraph} also analyse currently available AI policy frameworks, from industry, governments, general public, civil societies and academic bodies. Table~\ref{tbl:frameworks} includes many of the AI policy frameworks considered by \cite{ethicalGraph}.  They identify 47 different objectives, which they categorise into eight qualities. These eight qualities are: (1) privacy, (2) accountability, (3) safety \& security, (4) transparency \& explainability, (5) fairness \& non-discrimination, (6) human control of technology, (7) professional responsibility and (8) promotion of human values. We will map these eight qualities on technology properties in the next section.    

\subsection{FEAS: Fairness, Explainability, Auditability and Safety}
\label{ss:FEAS}
To build AI-based systems that fulfil the above-mentioned qualities distilled from policy framework, we argue that we should consider technologies for four properties: Fairness, Explainability, Auditability and Safety (FEAS). 

\begin{itemize}
    \item {\bf Fairness Technologies:\ } technologies focused on detection, prevention and mitigation of discrimination and bias 
    \item {\bf Explainability Technologies:\ } technologies focused on explaining and interpreting the outcomes and actions of AI-based systems to stakeholders~(including end-users) 
    \item {\bf Auditability Technologies:\ } technologies focused on enabling third parties and regulators to supervise, challenge or monitor the operation of the model(s) 
    \item {\bf Safety Technologies:\ } technologies focused on ensuring the operation of the AI-based system as intended in presence of accidental failures and malicious attackers
\end{itemize}

\begin{table}[]
\caption{Technology properties to establish the policy framework qualities \cite{socialTrustTechnology}.  \yes = technology property impacts the quality, \bigyes = technology property is pivotal to the quality}
\begin{tabular}{l| c | c | c | c |}
 \multicolumn{1}{c}{} & \multicolumn{4}{c} {{\bf Technology Property}} \\ 
{\bf Policy Framework Quality} & Fairness & Explainability  & Auditability  & Safety \& Security \\ \midrule
Privacy   &       &     & \yes          & \bigyes     \\ \midrule
Accountability &      & \yes            & \bigyes          &     \\ \midrule
Safety \& Security &       &             &      & \bigyes    \\ \midrule
Transparency \& Explainability &       & \bigyes            & \yes          &     \\ \midrule
Fairness \& Non-Discrimination & \bigyes      &            &         &    \\ \midrule
Human Control of Technology &      & \yes            &          &     \\ \midrule
Professional Responsibility & & & & \\  \midrule
Promotion of Human Values & \bigyes      &    &     &     \\ \bottomrule
\end{tabular}
\label{t:qualities}
\end{table}

Table \ref{t:qualities} establishes how the eight qualities derived in \cite{ethicalGraph} for policy frameworks relate to the FEAS properties.  In the table, \bigyes indicates the technology property is necessary in achieving the quality, and, vice versa, the quality sets requirements for the technologies.  \yes indicates the technology property impacts or is impacted by the quality, but not as strong or as comprehensive.  If there is no mark, the technology property is not typically necessary and less directly dependent on the requirements of the specific quality.  We discuss each of the qualities (as defined in \cite{ethicalGraph}) and explain the relation with the technology properties depicted in Table \ref{t:qualities}. 

{\em Privacy} in \cite{ethicalGraph} relates to control over the use of data, the right to restrict processing, etc. Within computing, this usually fits under security, an element of safety. Auditability usually comes with privacy implications, usually negatively impacting on privacy.  {\em Accountability} concerns elements such as assessing impact, legal responsibility, the ability to appeal, creation of a monitoring body, etc.  If a third-party is involved, this directly relates to auditability (\bigyes), and some require the support from technologies for explainability (\yes).  {\em Safety \& Security} is self-explanatory, as are {\em Transparency \& Explainability} and {\em Fairness \& non-Discrimination}, where we note that {\em Transparency \& Explainability} may require and impact technologies developed for auditability.  {\em Human Control of Technology} includes the ability to opt out and the human review of automated decisions, which primarily relates to technologies for explainability. {\em Professional Responsibility} refers to ethical design, scientific integrity, which, although relevant for each, does not strongly relate to any of the technology properties specifically.  Finally, {\em Promotion of Human Values} has many dimensions, but many of these are not directly impacted by technology properties. As marked in the table, the exception is the machine learning approaches for fairness that have emerged in recent years.  

The mapping in Table \ref{t:qualities} between qualities in policy frameworks and concrete technology classes always allows room for debate.  To test the mapping is reasonable, we went through each of the policy framework of Table \ref{tbl:frameworks} to identify the technology properties that relate strongly with the respective policy frameworks. Table \ref{tbl:frameworks} marks frameworks that refer to fairness, explainability, safety and auditability technology properties, using the symbols \yes and \bigyes as explained in the table caption.  As one can see from Table \ref{tbl:frameworks}, all four of the FEAS technology properties tend to be mentioned in each of the policy frameworks, with an emphasis on most frequently on Safety \& Security and Fairness.

\section{Trustworthy Machine Learning Life Cycle}
\label{s:chain}
\begin{figure}
	\centering	
	\includegraphics[scale=0.3]{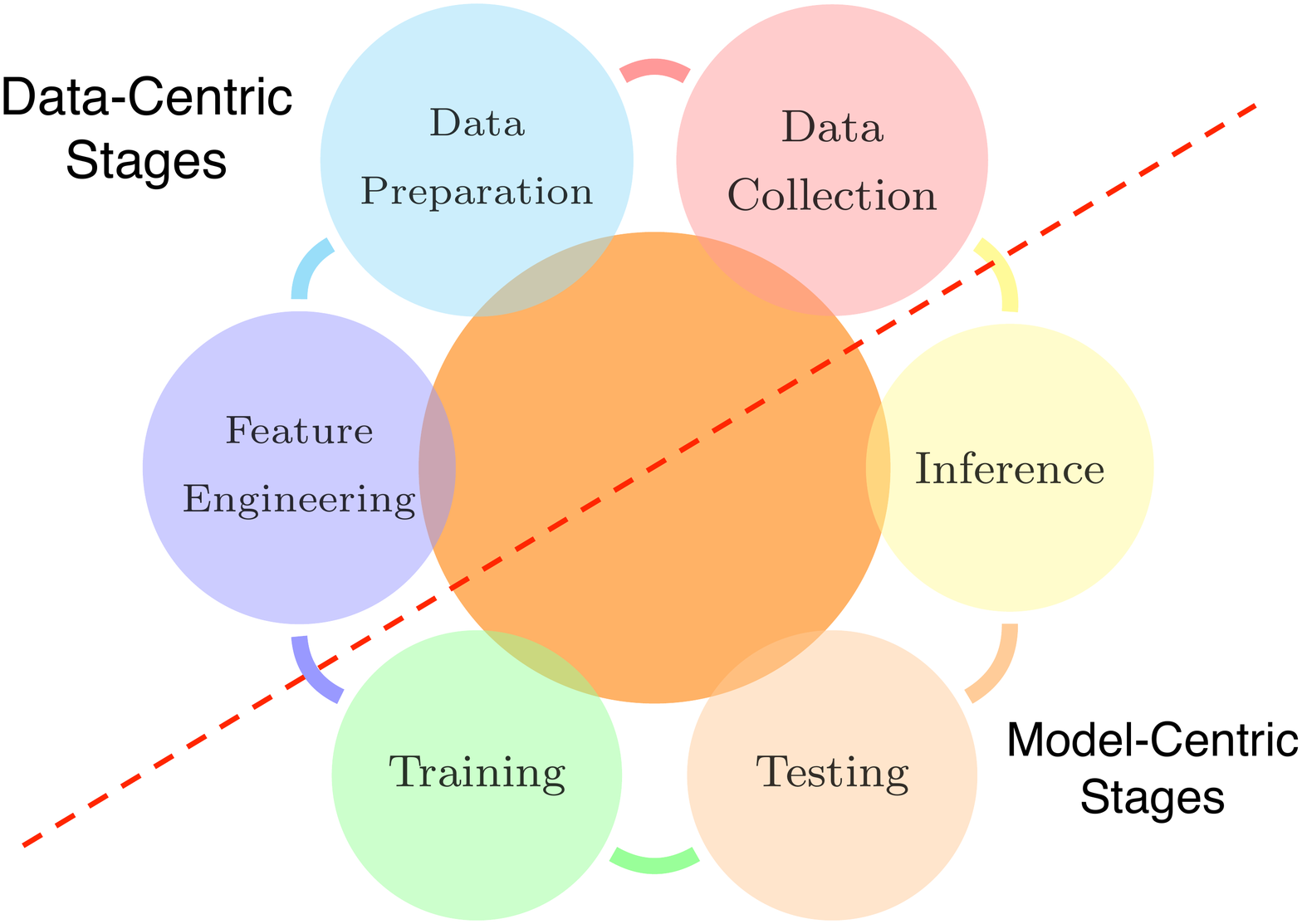}
	\caption{Various stages in the machine learning lifecycle, motivating the notion of a chain of trust.}
	\label{fig:pipeline}
\end{figure}

The literature presents life cycles for AI-based systems at vastly varying levels of abstractions. In this work, we use the chain of trust for machine learning (Figure \ref{fig:pipeline}), which we introduced in \cite{toreini2020relationship}. The chain of trust is inspired on the machine learning software pipelines proposed for real-life big data systems~\cite{polyzotis2018data}.  The pipelines combines in a single workflow software packages and tools that perform tasks such as data collection, streaming, analysis, visualization. Examples application for such pipelines include medical imaging \cite{gibson2018niftynet}, IoT \cite{qian2019orchestrating} or other production systems \cite{polyzotis2018data}.  

The stages identified in pipelines typically encompass the stages identified in Figure \ref{fig:pipeline}, worked out in detail for a particular domain. The precise stages that one naturally distinguishes within a life cycle depends on the type of machine learning that is applied. AI applications in robotics are considerably different from the big data oriented systems, systems based on reinforcement learning differ from these based on (un-)supervised learning, and a system that relies on online learning is inherently different from those using a batched learning approach (see, e.g., \cite{wachenfeld2016autonomous} for a discussion). We believe that the stages in the chain of trust are generic enough to usefully frame the discussion of trustworthy machine learning across the variety of AI-based systems that exists. 

The chain of trust in Figure \ref{fig:pipeline} contains two groups of stages, the {\em data-centric} stages and the {\em model-centric} stages. In Section \ref{s:data} and \ref{s:model} we discuss trustworthiness technologies for these two groups of stages, respectively.  

The data-centric stages consist of {\em Data Collection}, {\em Data Preparation} and {\em Feature Engineering}.  To establish services that fulfil the qualities proposed in the policy frameworks, it is critical to consider elements such as bias and privacy starting from the initial Data Collection. The data will go through other pre--processing to facilitate the application of machine learning techniques. This is represented by the stages Data Preparation, to `clean up' the data sets to be fit for the application of machine learning approaches, and Feature Extraction, in which we include feature engineering by experts, dimensionality reduction and feature selection. We would not include in this data-centric stage feature learning approaches such as principal component analysis or autoencoders, since these are machine learning techniques in their own right.   

The model-centric stages we distinguish are Training, Testing and Inference.  {\em Training} refers to the building of the model based on a subset of the data, the training data.  The {\em Testing} stage refers to evaluating the model arrived at in training, based on some metric. For example, in classification, a typical accuracy metric is the proportion of the test data that has been mapped correctly to their original label.  The {\em Inference} stage refers to the actual use of the model to make decisions or predictions.  Delineating between these three stages is natural in batched learning, in which quality data is available to train and test offline. In online or incremental learning approaches, e.g., used if the training data is too large to allow offline training or if the characteristics of the data change relatively rapidly over time, the stages cannot be delineated so clearly since training, testing and inference is intertwined~\cite{mohri2018foundations}.  

The different stages in the chain of trust are not independent. For instance, bias in the results of models are impacted by how the data was pre--processed. In Figure \ref{fig:pipeline} we represent this dependency between the stages through the chain. Moreover, systems will often iterate through (some of) the stages during their lifecycle, particularly in online learning settings where a model is continuously updated. The chain of trust is therefore represented a cycle in Figure \ref{fig:pipeline}.  We refer to \cite{toreini2020relationship} for a detailed discussion of various additional characteristics of the chain of trust.  

\section{FEAS: Definitions, Challenges and Assessment}
\label{s:challenges}
In this section we introduce the FEAS properties, providing definitions and discuss how to assess each property. 

\subsection{Fairness}
\label{ss:fairness}

A decision rule is said to be \emph{unfair} if it treats individuals with similar decision-related features in a different way due to personal characteristics that should not affect the final decision. Recent years have seen several examples of fairness issues in machine learning models~\cite{mehrabi2019survey}, for instance Amazon Inc. adopted a ML algorithm for its recruiting system that proved to be biased against women~\cite{dastin2018amazon}. As a consequence, fairness has has gained dramatically increased attention in both academia and industry, aiming to establish variants of fairness-aware machine learning, which  Friedler~et~al.~\cite{comparitiveFairness} defines as ``algorithms to provide methods under which the predicted outcome of a classifier operating on data about people is fair or non-discriminatory''.  The literature is extensive, crosses multiple disciplines, and arguable, as a new and fast moving field, still dispersed \cite{interdisciplinary}. 

Fairness is particularly important if it concerns discrimination in relation to data features about \emph{Protected Attributes} (PAs). These usually include (but are not limited to) ethnicity, gender, age, nationality, religion and socio-economic group. PAs are also referred to as \emph{sensitive} attributes in the literature.  California Law~\cite{disparateLaw} distinguishes disparate treatment and disparate impact. Disparate treatment occurs if the PAs affect the process output, while disparate impact happens when the output of the process disproportionately impacts depending on PAs. 

Machine learning algorithms are typically not inherently fair or unfair; however, the outcome of a machine learning model \emph{can} be unfair due to several reasons~\cite{mehrabi2019survey}. Collected data is, by its nature, a reflection of the real world. Thus, unfairness or discrimination in the real world would be reflected in related datasets, which would feed into the algorithms and therefore produce unfair outcomes. A dataset doesn't become fair by simply ignoring the protected attributes \cite{statiticalFramework}, since correlations between features could still lead to discrimination. For example, an individual's post code, degree, hobbies, etc., are likely to be correlated to their ethnicity.  In addition, other phenomena may contribute to bias in the date, for instance the lower frequency in data of underrepresented groups~(known as \emph{size disparity}), which may cause an inferior predictive performance for these minority groups. 

Designing a fair machine learning system requires being able to measure and assess fairness. Researchers have worked on formalising fairness for a long time. Narayanan \cite{narayanan2018translation} lists at least 21 different fairness definitions in the literature and this number is growing.  A full list of fairness definition can be found in~\cite{fairnessDefintions}. For the following definitions, let the set $A$ denote the protected attributes of an individual $U$, and $X$ the rest of attributes that can be observed by the algorithm. Let $Y$ be the actual label of an individual and $\hat{Y}$ the predicted label stemming from the machine learning algorithm. The different fairness definitions may be categorised as suggested by Kusner~et~al.~\cite{counterfactual}, as follows.  

\begin{itemize}

  \item{\bf Fairness Through Unawareness (FTU).\ } An algorithm satisfies FTU when its predictions are independent of the PAs~\cite{counterfactual}:
  \[
    \hat{Y} : X \rightarrow Y.
  \]
  The above definition simply discards the protected attributes~($A$) in the decision making process. FTU is also referred to as {\bf anti-classification}~\cite{corbett2018measure} in the literature. 

  \item{\bf Individual Fairness (IF.\ } An algorithm satisfies IF when similar samples yield similar outcomes. Generally, given a metric $d$ and some value $\delta$, if $d(U_i), U_j) < \delta$, the predictions for $U_i$ and $U_j$ should be roughly similar:
  \[
    \hat{Y} \left( X_i, A_i \right) \approx \hat{Y} \left( X_j, A_j \right).
  \]
  The problem with IF is that metric $d$ needs to be adequately defined (or learnt, see~\cite{counterfactual}). Thus, the implementation of such fairness is prone to increasing unfairness when not chosen correctly. 

  Joseph~et~al.~\cite{Joseph2016a} propose a metric known as \emph{Rawlisian Fairness} based on Rawl's notion of fair equality for opportunity. This notion emphasises that the individuals with same talents and motivations should have the same prospects of success. 

  \item{\bf Group Fairness.\ } This refers to a family of definitions, all of which consider the performance of a model on the population groups level. Most of the fairness definitions in this group are focused on keeping positive ratios proportional across groups. The definitions in this category are relevant to the disparate treatment and impact notions. The most popular group fairness definitions---as presented in~\cite{demographic, equalizedOdds}---follow, where $A=0$ and $A=1$ determine membership of a protected group or not.

  \begin{itemize}
    \item {\bf Demographic Parity (DP).\ } A classifier satisfies DP when outcomes are equal across groups:
      \[
        P \left( \hat{Y} = 1 \mid A = 0 \right) = P \left( \hat{Y} = 1 \mid A = 1 \right).
      \]
    \item {\bf Equalised Odds (EO).\ } A classifier satisfies EO if equality of outcomes happens across both groups and true labels:
      \[
        P \left( \hat{Y} = 1 \mid A = 0, Y = \gamma \right) =
        P \left( \hat{Y} = 1 \mid A = 1, Y = \gamma \right),
      \]
      where $\gamma \in \left\{  0, 1 \right\}$.
    \item {\bf Equality of Opportunity.\ } is similar to EO, but only requires equal outcomes across subgroups for true positives:
      \[
        P \left( \hat{Y} = 1 \mid A = 0, Y = 1 \right) =
        P \left( \hat{Y} = 1 \mid A = 1, Y = 1 \right).
      \]
  \end{itemize}

  We note that \cite{equalizedOdds} relates these metrics to disparate treatment and impact discussed above.  Other metrics of fairness related to disparate treatment may be found in~\cite{chouldechova2017fair, corbett2017algorithmic}.
          
  \item{\bf Counterfactual Fairness (CF).\ } The use of causal models can be a very powerful fairness analysis tool~\cite{counterfactual, kusner2020long}. Given a causal model, CF compares the probabilities of the possible outcomes for different \emph{interventions}---substituting the structural equations of a PA with a fixed value. Formally, a predictor $\hat{Y}$ satisfies CF if under any context $X = x$ and PA value $A = a$,
  \[
    P \left( \hat{Y}_{A \leftarrow a} = y \mid X = x, A = a \right) =
    P \left( \hat{Y}_{A \leftarrow a'} = y \mid X = x, A = a \right)
  \]
  for all $y$ and for any value $a'$ attainable by $A$.
\end{itemize}


The above definitions of fairness are statistical definitions that can be directly applied to machine learning. Grgic-Hlaca~et~al.~\cite{processFairness} argue that is not always possible or desirable.  They propose a procedure to ensure a fair use of attributes through users' feedback and call this {\bf Process Fairness (PF)\ }. An algorithm satisfies PF if the use of attributes is considered fair in the model assessed through some other means. 
  
It is important to note that establishing fairness with respect to one definition does not necessarily imply into fair output in another notion. in fact, it is theoretically not possible to fulfil all of the definitions at the same time for a model, as shown by Friedler et al.~\cite{fairnessImpossibility}.  Similarly, fairness definition are not necessarily easy to apply correctly, as demonstrated by Kusner et al.~\cite{counterfactual} who show that some specific definitions of fairness can increase discrimination if applied incorrectly, while \cite{fairnessImpossibility} show that slight changes in the training set has noticeable impact on the discrimination performance in the output, thus leading to result instability.  Clearly, designing discrimination-free models is not straightforward, something we will elaborate on when discussing open problems.

\subsection{Explainability}
\label{ss:explainability}

In the machine learning literature, the terms  \emph{explainability}, \emph{transparency}, \emph{intelligibility}, \emph{comprehensibility} and \emph{interpretability} are often used interchangeably~\cite{explainableSurvey}. Lipton~\cite{lipton2016mythos} provides a detailed discussion on terminology. 
The challenge in explainability is to bridge the gap between the operations of the algorithms and human understanding of how the outcomes has been obtained. According to Guidotti et al.~\cite{explainableSurvey}, explanation is an interface between the decision maker and humans. An explanation should strive to be \emph{accurate}, as it portrays the inner functionality of the machine learning model, as well as \emph{comprehensible}, as it intends to convey the decision process to humans. Herman~\cite{herman2017promise} sees this as a trade-off, between explaining a model's decision faithfully and presenting it a human understandable way. Guidotti et al.~\cite{explainableSurvey} suggest an explainable model must have at least two qualities in addition to accuracy, namely interpretability, or the extent to which a consumer of the explanation finds it understandable, and fidelity. The latter refers to the extent that an explainable model operates similar to a black--box model. Ideally, the explainable model must give the same results as the black--box model.

Doshi-Velez et al.~\cite{doshi2017towards} introduce two types of explainability, namely explainability with reference to a specific application, for instance a physician being able to understand the reasoning behind a classifier's medical diagnostic, and domain-independent explainability, where it is possible to understand how the classifier has produced one particular prediction. They describe three dimensions for such system: \textit{level} of the explanation, i.e., global vs local \textit{time} limitations, and  \textit{user expertise}.
Time limitations refer to the amount of time that the user needs to understand the explanation. In emergencies, such as surgical procedures and disasters, the users require a simple, straightforward explanation so they would spend minimum amount of time to understand the decision process. However, in less time--constrained situations, the users might prefer detailed explanations over simplistic ones. Also, the method of explanations is dependent on the user's level of expertise. For instance, a data scientist is more interested to know more about the inner patterns and parameters of the model and the data whereas an executive manager requires a bird-eye explanation of how a specific decision is determined~\cite{choo2018visual}. 

Technical approaches to providing explanations from predictive machine learning models can be {\em model-specific} or {\em model-agnostic}. Examples of model-specific explanations are discussed e.g. in ~\cite{freitas2014comprehensible}. These include not only linear regression and logistic models, which are arguably the simplest ``native explainable'' models, but also decision trees, classifiers such as SVM (Support Vector Machines), and Bayesian probabilistic classifiers. Model-specific explanations do not capture the complexity of highly non-linear models, including Neural Networks and Deep Networks in particular. 

Model-agnostic explanation approaches address this category of models in a \textit{post hoc} fashion, namely by accepting that the model is a \textit{black box}, and attempting to approximate its behaviour by repeatedly requesting predictions in specific areas of the feature space~\cite{Wachter2017}. A notable example of this approach is Lime~ \cite{Ribeiro2016a}, which seeks to provide local explanations in the form of linear approximations of the model that are accurate in small regions of the space. This is useful, for example, to explain why a certain individual has been denied a mortgage application. Other post-hoc model-agnostic approaches provide explanations in the form of a ranking of features, even when the underlying model is not-linear. These include, amongst others, InterpretML \cite{Lakkaraju2017,Nori2019} and Shapley Values~\cite{lundberg2017unified}. The main characteristic of both these approaches is to provide not only a ranking of features in order of importance, but at a more granular level, an indication of the importance of the feature as a \textit{function of its value}. For instance, an explanation may indicate that age becomes a more a dominant feature to predict a negative outcome for some disease as it increases, but it may be irrelevant for, say, young patients. Finally, Van der Schaar et al.~\cite{NIPS2019_9308} propose a breakthrough approach to model-agnostic explainability based on the notion that arbitrarily non-linear models can be approximated using a set of equations, from which explanations can be more easily derived.

There is consensus over the lack of evaluation framework to {\em assess} explainability of a model~\cite{miller2018explanation,doshi2017towards,schmidt2019quantifying,lipton2016mythos}. Kononenko et al.~\cite{kononenko2010efficient} suggest a qualitative approach to measure explainability. Letham et al.~\cite{letham2015interpretable} propose an evaluation framework that were limited in terms of scalability or needed prior expertise before using them. Other proposals~\cite{lundberg2017unified,samek2016evaluating}  propose frameworks that involve an explainability dimension along with other parameters. They mainly focus on the effects explainability has on computational efficiency on the model. Finally, Schmidt etal.~\cite{schmidt2019quantifying} propose an inclusive quantitative framework to measure explainability in various machine learning algorithms. Their proposed framework measures the effectiveness of the explanation for algorithmic decisions by focusing on the information transfer rate at which humans make the same decisions. They also defined a trust metric which determines how humans are willing to accept the algorithmic explanations.

\subsection{Auditability} 
\label{sec:auditability}

Auditability refers to the methods that enable {\em third parties} to verify the operations and outcome of a model. The motivation for auditing very often stems from the other three properties, fairness, explainability and safety \& security.  For instance, since model fairness is affected by bias in the training set as discussed in Sec.~\ref{ss:model:fair}, one may want to establish an auditable trail to decide whether the bias was native to the raw data, and whether it was affected by any of the operations performed during data pre-processing. 

Auditibility of the model generation process and the model operation is important for the implementation of regulations such as the European GDPR legislation. At its core, GDPR is about data protection, posing requirements on industry, government and other data owners about safeguarding the date, transparency of the data collection process and explicit consent from an individual for the use of its data.  This implies that under GDPR people have the right to challenge the data collection, have control over their collected data and maintains the ``right to be forgotten'' from the data source.  \cite{perreault2015big} notes a trade-off between keeping data private and the ability of machine-learning systems to perform certain tasks.  This trade-off is a key concern behind policy shaping for AI-based systems.  

Particularly interesting is that GDPR also requires ``suitable explanations'' to be provided for any decision made by an algorithm~\cite{Goodman2016}. Providing mechanisms to audit a model, as well as the process used to obtain a model, ensure that the model itself is accountable for the decisions it produces. The case for accountability in algorithmic decision making has also emerged outside GDPR, most notably in~\cite{10.1145/2844110} and in other countries such as India~\cite{burman2019will} and China~\cite{sacks2018new} legislation is under review.  In turn the need for accountability  entails validating the fairness of a model.

\subsection{Safety \& Security}
\label{sss:safeDesign}
The category of safety and security properties refers to the robust operation of systems in real-life operation.  This covers a wide variety of concerns, from safe vehicles to protection of data against privacy attacks. In this paper we emphasise the review of literature with respect to security and privacy.   For a discussion of terminology and a broad perspective on the design of dependable systems, see \cite{Avizienis04}.  

Security of AI-based systems and services refers to the protection of these systems against malicious attacks. Traditionally, the objective in securing a system is expressed through the CIA triad: Confidentiality, Integrity and Availability \cite{stamp2011information}. \emph{Confidentiality} attacks are designed to expose the structure, configuration of a model or privacy leakage to access the data. Privacy equates to maintaining confidentiality.  Attacks to the \emph{integrity} attempt to alter the output of model to the adversary's choice or change the behaviour of the model to the attacker's intention. Finally, \emph{availability} attacks aim to prevent the model's usual service to the client or getting meaningful results.

A perfectly secure system does not exist \cite[p.~21]{anderson2008security} and this holds also for AI-based systems \cite{barreno2006can}. The security of a machine learning system (see Section~\ref{ss:model:safety}) are impacted by the chosen cryptographic primitives, hardware vulnerabilities, protocol inefficiencies in design and implementation as well as the behaviours of humans implementing, deploying and using it. Clearly, the current large-scale adoption of machine learning motivates attackers to actively engage into detecting design weaknesses or implementation bugs and leveraging them to collect information~\cite{szegedy2013intriguing}, modify data~\cite{biggio2014poisoning,testingAttack4,biggio2011support,biggio20142}, contaminate the operation of the algorithm~\cite{kloft2010online,sharif2016accessorize,barreno2006can,szegedy2013intriguing}, exploit leakage of sensitive information~\cite{drucker1999support,krizhevsky2009learning,erhan2010does,trainingAttack1,lowd2005good} and much more~\cite{narodytska2017simple,sharif2016accessorize,carlini2016hidden,nguyen2015deep,testingAttack5,papernot2016limitations,liu2016delving,kurakin2016adversarial}.  

Understanding the attacker is the building block of any defensive mechanisms~\cite[p.~20]{anderson2008security}. Attacks on machine learning systems exhibit intention and capability \cite{trainingAttack5} or, similarly, goal, knowledge, capability and attacking strategy, see Baggio et al.~\cite{securityassessment}.  In ~\cite{securityassessment}, the {\em goal} of the attack describes the desired impact of the attack in accordance with the CIA model. The goal's impact can be of two types: targeted attacks, which aim to damage a specific victim, or indiscriminate attacks, which aim for extensive damage with no exception. {\em Knowledge} determines the degree of intelligence that the attacker has at the time of the execution of the attack. A black-box adversary designs the attack without any knowledge of the model`s internal configurations, whereas a white-box adversary knows the settings in which the model is trained and operates. In the real-world, the black-box attacks is reportedly more common~\cite{MLSecuritySurvey}.  {\em Capability} determines the degree to which the attacker is able to successfully implement an attack scenario and get access to the data or the model. Capability can be a qualitative feature with regards to three aspects~\cite{securitySurvey}: whether the security threat is causative~(i.e. direct manipulation of the model or data) or exploratory~(i.e. injection of manipulated data to mislead the model), the percentage of the training or test data that is controlled by the attacker and to which extent the attacker knows the model configurations and its features. Capabilities increase significantly if one considers insiders, that is, individuals that know about one or more of the pipeline stages. To avoid weakening the overall system ~\cite{MLSecuritySurvey} argues that there should be different levels of access to different actors involved in the model pipeline, i.e. \emph{data-owners, system providers, consumers and outsiders}. The {\em attacking strategy} deals with the analysis of the behaviour of the attacker at the time of the exploit. There are usually different ways to implement an attack with a certain goal. Strategy refers to selection of the optimal function that characterises the goal.

Security assessment of a system is usually based on a ``what-if'' analysis~\cite{securitySurvey}. The appropriate defensive mechanism is implemented based on the conditions of the data, model and the environment that the system is going to be deployed. The designers should provide an assessment of the vulnerabilities of the model, too~\cite{quantitativeSecurity2}.

When designing a machine learning system for security, one needs to combine a reactive design approach with a proactive approach~\cite{securitySurvey}. In  reactive design, the designer implements defensive mechanisms after a vulnerability is detected. However, the current pace of progress in machine learning and applications justifies one to consider security and privacy from the initial stages of designing a model until the model operates in the wild. Thus, the trend is to take a proactive attitude and anticipate common threats through a set of \emph{what--if} queries~\cite{rizzi2009if}. For instance, in a proactive design approach, the model designer may invite a red team to attack the system before deployment. In either approach one needs to realistic analysis of the potential attackers for the goal, knowledge, capability and strategy dimensions of attackers identified by \cite{securityassessment}.

Both approaches take an empirical way to tackle security. Thus, they rely on a predefined set of \emph{known} attack scenarios. As a result, they lack comprehensive vision for a safe system because they do not take into account creative and novel attacks in their threat model. The more vital applications of machine learning require stricter approach in the design by modelling the behaviour of the system mathematically~\cite{wu2016methodology,gajane2017formalizing,kuznetsov2001machine,kuznetsov2004machine,li2017formal,shalin1988formal}. 



\section{Data-centric Trustworthiness Technologies}
\label{s:data}
In this section, we review trustworthiness solutions in the data--centric stages in machine learning pipeline (see Figure \ref{fig:pipeline}), that is, data collection, pre--processing and feature selection, respectively. The subsequent sections review each of FEAS properties.  



\subsection{Fairness in the Data-Centric Stages}
\label{ss:data:fair}
Data inevitably reflects the nature of the environment that it has been obtained from as well as the decisions made in that environment. Pedreshi~et~al.~\cite{pedreshi2008discrimination} aptly state that ``learning from historical data may lean to discover the traditional prejudices that are endemic in reality''. 

An important class of bias in ML is \emph{sample bias}~\cite{dataScienceView}. This refers to sampling in such a way that not all PA groups are adequately represented, i.e. to the dataset being \emph{imbalanced}~\cite{lemaitre2017imbalanced}. This kind of bias is not to be confused with \emph{statistical bias}, which refers to a systematic error in the expected result of a process, differing significantly from the true value of the quantity being estimated~\cite{dietterich1995machine}. The notion of sample bias is in direct relation with group fairness definitions as explained in section~\ref{ss:fairness} since they both concern with maintaining a~\emph{balanced} ratio of data vs. output in the machine learning system. There are two possible sampling bias directions for PA groups, namely to be \emph{over} or {under} represented~\cite{dataScienceView}. Naturally, one PA group being overrepresented means that some of the remaining groups will be underrepresented. This poses the risk of a learning algorithm being giving up on accuracy for the minoritary groups for improved overall accuracy~\cite{lemaitre2017imbalanced, disparateLaw, fairclassifier}. An extreme case of underrepresentation would be the existence of a group being unknown to the algorithm. 

A first step to reduce bias is to ensure the data has been collected in an adequate manner~\cite{altmann1974observational}. This implies that the collection process should yield representative data from all PA groups. This is not always easy to achieve, since full understanding of the environment and full control over the data-collection processes are needed. In order to prevent sampling bias, a possible approach is \emph{stratified sampling} methods~\cite{neyman1992two}.

A potential strategy to avoid bias is to learn about the unintended discrimination that may occur in the specific problem one is  resolving. Such \emph{awareness} is vital but not sufficient to prevent discrimination~\cite{dataScienceView}. The active learning approach involves \emph{expert} human supervision in the collection to improve the process~\cite{settles2009active}. Alessandro~et~al.~\cite{dataScienceView} suggest that data scientists should also test and document all the features that have high correlation with the minority groups. Also, data scientists should perform a statistical demographic analysis to gain knowledge about groups that are prone to discrimination.

A different approach consists of pre-processing the data, with the aim of feeding an unbiased training set into the model~\cite{lahiri2013resampling,efron1994missing}. These pre-processing methods can be based on either class labels or on attributes. In the class labels approach, data is relabelled to reduce disparity; this is sometimes referred to as \emph{massaging} the dataset. These correction methods may be generic~\cite{Kamiran2012}, or adjusted with respect to a particular combination of \emph{world-views}~\cite{friedler2016possibility} and fairness definitions~\cite{zelaya2019parametrised}. Alternatively, Luong~et~al.~\cite{luong2011k} assign a decision value to each data sample and then adjusts this value to eliminate discrimination. 

In the {\em attributes approach} to pre-processing, one either maps the attributes into a transformed space~\cite{calmon2017optimized} or combines the values of the protected attributes with the  non-protected \cite{demographic}. Feldman~et~al.~\cite{demographic} propose to modify each attribute so that marginal distribution of the PA and non-PA values remain equal all over the dataset.

Pre-processing techniques have no awareness about the internal operation of a machine learning model and as a result they are not tuned for the the optimal functioning of the model, resulting in a trade-off between fairness and accuracy~\cite{zliobaite2015relation}. This, however, makes most of them model-agnostic and highly interpretable. Relabelling of the training data, though, potentially affects the explainability of the model~\cite{zafar2019discrimination} and has a steeper fairness-accuracy trade-off~\cite{zelaya2019parametrised}.

\subsection{Explainability in the Data-Centric Stages}
\label{ss:data:explain}
Data has different degrees of explainability to humans~\cite{explainableSurvey}. 
The vast majority of machine learning algorithms use data representation in vector and metrics forms~\cite{explainableSurvey}. Johansson et al.~\cite{huysmans2011empirical} suggest that \emph{tables} are the most understandable data format for humans and algorithms. Tables are a familiar format of data representation for humans, while the process of converting them to vectors and metrics is straightforward. \emph{Images} and \emph{texts} are also understandable to humans but conversion to matrices and vectors and back is not straightforward, thus making it more challenging to explain the steps of the algorithm ~\cite{explainableSurvey}..  Explainable algorithms that can process images and texts would require a transformation processes such as equivalences, approximations or heuristics for explaining the decision making process. 

Explainability in other types of data such as sequence data, spatio--temporal data and complex network data, has not yet been researched. A potentially effective explainability tool is to visualise the data (and the steps of the algorithms). Visualization could assist in identifying correlations and connections between seemingly unrelated variables. In practice, however, the high dimension of the data does not allow comprehensive visualisation. One solution is transforming high-dimensional data into a two or three-dimensional space, e.g., ~\cite{maaten2008visualizing}.

Zelaya~\cite{zelaya2019volatility} highlights the impact of the appropriate data--processing methodology before training of the model in the overall explainability of the machine learning system. This includes the data-centric stages of data collection, pre-processing and feature selection. Zelaya~\cite{zelaya2019volatility} defines a metric which compares the outputs of classifiers trained over two training sets, an unprocessed and a pre--processed one. This comparison allows one to analyse the impact of pre-processing on the explainability of the model. Blum et al.~\cite{Blum1997} present a survey on data selection methods for large datasets containing redundant and irrelevant data samples. They focus on two key issues in dataset, first, selection of features and second, selection of most useful samples from the dataset.  Crone et al.~\cite{Crone2006} study the effects of pre--processing methods such as scaling, sampling and encoding variables over prediction performances.  Similar analyses have been performed for text data such as documents~\cite{Goncalves2010}, text classification~\cite{Uysal2014} and sentiment analysis~\cite{duwairi2014study}. Other directions such as~ \cite{Cai2018}, followed the same approach.  

In our opinion, explainability in the data collection, pre-processign and feature engineering stages of the machine learning life cycle has been underexposed compared to work on algorithsm.  Arguable, even for explainable algorithms there is a clear need to explain the sources of data and how it was processed. Labelling standards in supervised machine learning are only a small step and the area deserves much more attention.  

\subsection{Auditability in the Data-Centric Stages}
\label{ss:data:audit}
Auditing of the operations of a real-life machine learning system inherently requires one the monitor across all the stages of the chain of trust, from data collection all the way to inference.  technology solutions therefore also tend to span all stages. This is particularly true for provenance solutions that target machine-learning systems ~\cite{lineageSurvey}. 

We therefore refer to Section \ref{ss:model:audit} for an extensive discussion of provenance technique for machine learning systems.  


\subsection{Safety \& Security in the Data-Centric Stages}
\label{ss:data:safety}
There exist a plethora of techniques for safe and secure information systems. In this section, we first discuss how these general concerns apply to systems and services that are based on machine-learning, with a particular emphasis on big data.  Then we discuss interesting recent developments in privacy through differential privacy and data-related attacks on machine learning associated with data poisoning. 


One of the main differentiators of many modern-day machine-learning systems is the large volume of data that requires to be stored or streamed. The sensitivity of the analysis implies that there is an emphasis to store the data securely to preserve privacy and security against various types of attacks is imminent. 
Perdosci et al.~\cite{perdisci2006misleading} managed to implement a real--world attack in Polygraph~\cite{newsome2005polygraph}~(a worm signature generation tool) by systematically injecting noise into the flow detection dataset and consequently reduce the accuracy of model intentionally.

Lafuente~\cite{lafuente2015big} suggests several security strategies for storing big data, both structured and unstructured. First, the data should be {\em anonymised}, meaning personal identifiable information is removed from the dataset. Note that this alone may not be sufficient for privacy protection since the anonymised data can be cross--referenced with other sources of the data to leak information, a process is known as de--anonymisation or linking attacks in the literature~\cite{narayanan2008robust}. Second, the data must be {\em encrypted}. This is particularly the case when it comes to cloud--based machine learning services~(i.e. Machine Learning as a Service~\cite{ribeiro2015mlaas,hesamifard2018privacy}), both for storage and data transmission.  Since the data requires processing, and thus decryption, encryption needs to be complemented by strong security guarantees of the (virtualized) servers. To make sure the data is only decrypted by the party that owns the data, one would utilize reliable key distribution or concepts such as attribute--based encryption~\cite{bethencourt2007ciphertext} which enables a fine--grained access control on the encrypted data. Third, access control and monitoring is essential to protect the data. Usually, a combination of cryptography and access control forms an appropriate approach~\cite{lafuente2015big}. However, even popular systems such as Hadoop lack a proper authentication solution~\cite{jam2014survey}. Real--time monitoring is typically accompanied by threat intelligence systems and up--to--date vulnerability detection. The fourth strategy around governance, including implementing regulatory frameworks such as GDPR (see also auditability). 


An important recent concept with respect to privacy is to store data using cryptographic primitives that provide {\em Differential Privacy} \cite{dwork2011differential}. In differential privacy, the calculation of the result by a machine learning algorithm is not sensitive to change in an individual data item. In other words, the output of a machine learning system does not reveal information about an individual. Differential privacy models have two main benefits in comparison with the conventional privacy preserving mechanisms. First, it protects against attackers with full knowledge of the data records except for the target record, i.e., the strongest possible attacker in practice. Second, differential privacy properties of functions are mathematically proven. Differential privacy is considered one of the main trends in the privacy-preserving machine learning literature~\cite{MLSecurity1,MLSecurity2,MLSecurity3,MLSecurity4,wang2014compressive}. 

To assure integrity of the data, pre--processing techniques benefit from tampering--prevention measurements~\cite{MLSecuritySurvey}. The integrity attacks in this category work based on polluting or poisoning the datasets. Poisoning attacks on the training data aim to alter the model that is derived during the training stage.  Poisoning attacks on the testing set attempt to manipulate the test data distribution in order to separate it from the training set. This would eventually increase the likelihood of false recognition in presence of polluted data when the machine learning model is operating in the inference stage~\cite{xu2017feature,feinman2017detecting,metzen2017detecting}.  

Security architects of a machine learning systems evaluate the degree of malicious modifications of data with probability $\beta$ in their security assumptions. Kearns et al.~\cite{kearns1993learning} determine an upper bound for $\beta$. In their analysis, a machine learning algorithm with an error rate of $\epsilon$, requires attackers to modify a fraction less than $\frac{\epsilon}{1+\epsilon}$ of the training set. The attacker with access merely to labels of the training data is constraint to a limited attack surface, requiring one to determine the most useful labels to manipulate~\cite{MLSecuritySurvey}. In this case, the attacker can operate by perturbing the labels randomly or heuristically. Baggio et al.~\cite{biggio2011support} demonstrate that to impact the efficiency of SVM classifier, random manipulation of 40~\% of labels in a training set is required.  In contrast, it requires 10~\% of the labels to be manipulated if it is done purposely. 

A number of further advances have been made with respect to data poisoning attacks. If attackers gain access to all the features in addition to the labels in training set, then they would be capable of implementing more serious attack scenarios~\cite{mei2015security,xiao2015feature}. They would benefit them the most if training stage is performed online~\cite{zhao2017efficient}~(i.e. the training dataset is updated regularly as the data is being collected). In online setting, Kloft et al.~\cite{kloft2010online} poison the training set of a clustering system for anomaly detection and the same approach has been applied to malware clustering techniques~\cite{biggio20142}. In offline learning, Biggio et al.~\cite{biggio2012poisoning} propose an attack that detect the target sample in the training set and degrades classification accuracy accordingly, assuming the classifier is a SVM. Xiao et al.Z\cite{xiao2015feature} propose the same strategy to degrade the functionality of feature selection algorithms like LASSO.

The countermeasures to malicious poisoning attack rely on monitoring the distribution characteristics of training and testing datasets. The existence of adversarial samples in a data set deforms the distribution; thus, some researchers propose anomaly detection techniques based on data distributions in adversarial environments~\cite{securityassessment,quantitativeSecurity2,fredrikson2014privacy,githubSherry}. Grosse et al.~\cite{grosse2017statistical} introduce two statistical metrics, Maximum Mean Discrepancy and Energy Distance, to measure the statistical properties of sample distributions especially on test samples. Their proposal successfully detects malicious injected samples in several data sets. 

Mitigation techniques for data integrity focus on ensuring the purity of training and testing data sets~\cite{securitySurvey}. Data sanitization is a dominant technique in this area, e.g., ~\cite{trainingAttack5} propose Reject on Negative Impact to protect spam filters against malicious poisonings. The basic idea of this method is to eliminate samples with large negative impact in the training set. They also propose an error estimation method in online training mode to compare the error rates of the original classifier and the one after the update. Laishram et al.~\cite{trainingAttack6} propose a solution protect SVM classifier from poisoning attacks. Finally, some proposals focus on feature reduction algorithms to prevent poisoning attacks. Rubinstein et al.~\cite{trainingAttack2} proposed an PCA--based poisoning detector based on techniques from robust statistics. Their method substantially reduces poisoning in various intrusion detection scenarios. 

\section{Model-centric Trustworthiness Technologies}
\label{s:model}
In this section we discuss technologies for the model-centric stages in the machine learning chain of trust, that is Training, Test and Inference stages.  The following sections survey of technologies with respect to each if the four FEAS properties.


\subsection{Fairness in the Model-Centric Stages}
\label{ss:model:fair}
The research in fairness for machine learning models is developing very fast.  In this section we aim to present a high-level perspective on types of approaches presented in the literature, with examples. To that end, we follow \cite{hajian2016algorithmic} and distinguish two types of model-centric solutions for fairness: \emph{in-processing} solutions aim to design an algorithm that is inherently fair~\cite{comparitiveFairness,madras2019fairness}, while \emph{post-processing} solutions adjust the outcomes so the outcomes are fair~\cite{hajian2014generalization}. Within in-processing approaches, one can distinguish between those modifying the model to remove unfairness and those in which an independent agent carries out the detection (or prevention) of unfairness.

Using the in-processing approach various solutions introduce a fairness-regulating term in the algorithm.  For instance, Kamishima~et~al.~\cite{fairclassifier} propose a regularization term for a logistic regression classifier: this term penalizes existing correlations between the protected attributes (PAs) and the predicted labels, leading to fairer classification. Kamiran~et~al.~\cite{kamiran2010discrimination} propose a fair decision-tree algorithm, which defines a criterion to either split or label a leaf node in order to remove discrimination. Other approaches add a set of constraints to the model and then optimise them simultaneously while the model is training. For instance, Goh~et~al.~\cite{goh2016satisfying} use a non-convex constraint to optimise a classifier under a fairness constraint based on disparate impact. Quadrianto~et~al.~\cite{quadrianto2017recycling} propose a fairness-definition agnostic unified framework. They tune two ML algorithms~(privileged learning and distribution matching) to enforce different notions of fairness. Woodworth~et~al.~\cite{woodworth2017learning} present a statistical method used in the learning process and propose a non-discrimination definition that increases the fairness of the learning algorithm. Zafar~et~al.~\cite{zafar2015fairness} introduce flexible fair classification and suggest a novel fair boundary decision threshold. They apply their approach to logistic regression and support vector machines.  And, as a last example of introducing fairness regulation in the algorithms, Calders~et~al.\cite{calders2010three} train Naive Bayes models for every possible PA value and then adjust the model to balance the outcomes. 

A second important in-processing approach is the use of \emph{adversarial learning} approaches. In adversarial learning, two models are simultaneously trained: the first one modelling the data distribution and the second one estimating the probability of a sample coming from the training data rather than from the model, the objective being to maximise the probability of the first model causing the second one to make a mistake~\cite{goodfellow2014generative}. Examples of adversarial learning for fairness are Beutel~et~al.~\cite{beutel2017data}---who aim at enforcing equality of opportunity---and Zhang~et~al~\cite{mitigatingAdverserialLearning}; this second method is applicable to multiple fairness definitions and gradient-based learning models. 

Post-processing approaches analyze the outcomes of a model and then adjust them to remove unfairness. An important limitation of these solutions is that they require access to PAs at the moment of decision making. Post-processing techniques can therefore not be used when there is no access to the PAs. There are various examples of post-processing approaches. For instance, Hardt~et~al.\cite{equalizedOdds} design a post-processing approach which adjusts a learned predictor in order to remove discrimination with respect to the equalised odds rate in the results. Corbett-Davis~et~al.~\cite{corbett2017algorithmic} work with three notions of fairness: demographic parity, conditional demographic parity---in which proportions across PA groups should remain constant given a limited set of legitimate risk factors, i. e. attributes naturally related to the classification task of interest---and predictive equality, which is satisfied when there are equal false positive rates across PA groups. They then introduce an algorithm to assign a discrimination score to a classifier, which in turn allows for detecting unfairness. 

In addition to the above specifically model-centric approaches, one can combine data-centric and model-centric solutions. Zemel~et~al.~\cite{individualFairness4} propose an algorithm optimising individual and group fairness at the same time through an adequate representation of the data, obfuscating membership to PA groups. They argue that there is a close connection between individual fairness and the differential privacy technique of hiding attributes of data through the addition of noise without it losing its utility~\cite{dwork2006calibrating}. Finally, Dwork~et~al.~\cite{decoupledFairness} propose a \emph{combined} in-processing and post-processing method based on training different classifiers for different PA groups.

\subsection{Explainability in the Model-Centric Stages}
\label{ss:model:explain}
Explainability of machine learning models has also experienced a surge in research, and as for fairness, our aim in this paper is to provide a high-level overview of the main types of explainability techniques, including examples. Therefore, we adopt the classification of Guidotti et al.~~\cite{explainableSurvey}, who distinguishes the following types model explainability approaches: explaining the outcome of the model, known as \emph{outcome explanation} and explaining the inner operation of the model, known as \emph{model explanation}. In addition, within model explanation, we distinguish explanation through \emph{transparent models} in which the model is designed for explainability, effectively interpreting its own inner complexities, as well as techniques for {\em representation of the model} based on documentation, visualization or provenance information.

{\em Outcome explanation} does not consider the inner functionality of models and is therefore also referred to as black box interpretability models~\cite{schmidt2019quantifying}. These solutions provide an interpretable model that explains prediction of the black box in human--understandable terms~\cite{explainableSurvey}. Outcome explainers are further classified into \emph{local} and \emph{global} explainers~\cite{lime}, where local explainers are designed to explain the outcome of a specific model and global explainers (also called agnostic explanator \cite{ribeiro2016model}) aim to be independent of the machine learning model type or the data \cite{explainableSurvey}. 

A number of approaches to global explainers have been proposed in the literature.  For instance, Local Interpretable Model-agnostic Explanations~(LIME) \cite{lime} matches the outcome of any classifier using the local approximation of the model's outcome with an explanation model. In their proposal, the authors define a desired explanation model and provide the outcome explanation in their desired criteria. \emph{Black Box Explanations through Transparent Approximations (BETA)}~\cite{Lakkaraju2017} proposes to explain the outcome of any classifier using a framework to quantify the unambiguity, fidelity and interpretability.  It generates explanations in the form of decision sets~(i.e. nested if-else statements which divide data into exclusive regions) and then optimizes with respect to a combination of the mentioned quantities. In this manner, BETA ensures the generated explanation is both understandable and comprehensive. 

Local explainers are designed for a specific machine learning models or class of machine learning models. Given the popularity of neural networks, it is not surprising that many local explainable solutions focus on neural networks.  For instance, \emph{Deep Learning Important FeaTures~(DeepLIFT)}~\cite{shrikumar2017learning} explains neural networks by assigning importance to every feature through back propagation. Building on developments such as \cite{shrikumar2017learning},  \emph{SHapley Additive Explanations (SHAP)}~\cite{lundberg2017unified}) proposes a unified framework for local explainers which generalises six existing methods, including DeepLIFT.  We point to Montavon et al.~\cite{montavon2018methods} for a more detailed tutorial on explainable neural network.

Local explainers have also been developed for other machine learning models that neural networks.  An approach for clustering algorithms is to represent each cluster with a prototypical example~\cite{kim2014bayesian}, which provides insights into the process that formed each cluster. Other local explainers have been developed for ensemble decision trees~\cite{chipman1998making,domingos1998knowledge,gibbons2013cad,zhou2016interpreting,schetinin2007confident,hara2016making,tan2016tree,deng2019interpreting,tolomei2017interpretable} and SVN models~\cite{nunez2002rule,fung2005rule,martens2007comprehensible}. 
{\em Model explanation} aims to explain the internal operation of the model instead of outcomes.  From a question about why the model yields specifics outcomes, the question becomes: \emph{why does the model behaves the way it does?}.  Obviously, the answer to this question must be expressed in a human--understandable manner. The model explanation approaches can be divided into those that use an explainer that mimics the behaviour of the model and the ones that aim to develop a model that can explain itself (transparent models). In model explanation usually the model behaviour is generated either as a {\em decision tree} or a series of {\em rule--based} conditions tailored for a specific type of machine learning model~\cite{explainableSurvey}. In addition, there exist solutions that follow an agnostic approach through which any type of model gets to be explained~\cite{explanation3,henelius2014peek,krishnan2017palm,lou2012intelligible,Lou2013}.

In the decision tree methods, a decision tree defines the conditions according to which the model decides. Such an approach allows one to represent an approximation of the black--box behaviour in a form which is understandable to humans. Craven et al.~\cite{craven1996extracting} propose a comprehensive explanation of the neural network by constructing a decision tree using a special set of queries that they called as Trepan, and a similar approach has been applied in~\cite{tan2018introduction}.  For neural networks, Krishnan et al.~\cite{krishnan1999extracting} and Johansson et al.~\cite{johansson2009evolving} use genetic programming to determine the decision tree and also for explanation of tree ensembles~(such as random forests) methods have been proposed to explain using decision trees~\cite{chipman1998making,domingos1998knowledge,gibbons2013cad,breiman2017classification,schetinin2007confident}. 

Rule--based approaches aim to explain the model through a \emph{set of rules} derived from the behaviour if the model. This has for instance been done for neural networks in Cravenan et al~\cite{craven1994using}. They propose a solution to map rule extraction~(a search problem) into a learning problem~\cite{explainableSurvey}. Other solutions in this category are~\cite{johansson2003rule,zhou2003extracting,augasta2012reverse}, including proposals explaining Support Vector Machines~\cite{fung2005rule,martens2007comprehensible} and tree ensembles~\cite{deng2019interpreting}.

The transparent model approach aims to make a specific model inherently explainable. As a general approach, one aims to express the impact of coefficients and features on the explainability of the model~\cite{rivest1987learning}. Also, ``Generalised additive models with interactions''~(\emph{GA2M}), proposed by Lou et al.~\cite{Lou2013}, is one of the more well--known transparent models. Impact can be expressed through different models, including single decision, decision trees and decision lists~(\cite{rivest1987learning}), in which the data is split in a threshold--based mechanism. Transparent models have been researched for different machine learning techniques, including linear models \cite{haufe2014interpretation} and neural networks~\cite{montavon2018methods,samek2019explainable,samek2016evaluating}.  In general, transparent models tend to be less accurate than their non-transparent counterparts, so considerable research is still needed in establishing methods that retain the accuracy levels while achieving explainability \cite{explainableSurvey}. 

Well thought-through representation of the data, the model, the outcomes and the accuracy is another way to improve explainability. Yang et al.~\cite{yang2018nutritional} introduce ``nutritional labels'', which are a collection of metrics on fairness, stability and explainability of individual ranking based systems such as credit scoring. They implement a web application that demonstrates these metrics as visual widgets. Mitchel et al.~\cite{mitchell2019model} suggest another documentation solution named ``model cards''. These cards are short documents accompanying trained machine learning models that include benchmarks across different populations, together with the kind of model, its target objective and intended use, as well as the applications that the model should not be used for. It also includes a list of the features that affect the model's outcomes, as well as a list of ethical considerations and usage recommendations. They propose model cards for visual and natural language processing. A related approach is to visualise the decision making process using \emph{provenance graphs}, a causality graph that is a directed cyclic graph, enriched with annotations capturing information pertaining to pipeline execution. In~\cite{Moreau2011}, the provenance graph of a model includes an annotated workflow of the methods of data collection, pre--processing techniques, the training process of the model and finally the parameters used for the model. This results in a complete picture of the machine learning pipeline. 

\subsection{Auditability in the Model-Centric Stages} 
\label{ss:model:audit}

We introduced model auditability earlier, in Sec.~\ref{sec:auditability}, where we briefly introduced techniques to achieve accountability in models.
Here we briefly review three kinds of approaches to ensure that the \textit{process} used to create a model is auditable.

Firstly, one may try to formally prove that a prediction is correct. This is the approach taken for instance by SafetyNets, a framework that enables an untrusted server (the cloud) to provide a client with a short mathematical proof of the correctness of inference tasks performed by a deep neural network~\cite{ghodsi2017safetynets}, with negligible impact on the accuracy and efficiency of the model.

A second and emerging strand of research is centred on \textit{data provenance}. This is ``information about entities, activities, and people involved in producing a piece of data or thing, which can be used to form assessments about its quality, reliability or trustworthiness.''\cite{w3c-prov-dm}. Here the idea is that provenance may be able to  provide a useful account of all data transformations, e.g. across a complex pipeline and further into the learning algorithm. Singh et al.~\cite{singh2018decision} suggest a legal perspective on  \textit{decision provenance}, which entails ``using provenance methods to provide information exposing decision pipelines''.

On more technical grounds, contributions concerning provenance focus on multiple aspects.  Firstly, one key concern is the reproducibility of the model-generation pipeline. For this, systems have been proposed that attempt to automatically capture some of the metadata pertaining to a model, such as the settings of the deep neural network \cite{schelter2017automatically}. Pipelines are typically implemented using bespoke scripts. The VAMSA tool aims to ``reverse engineer'' the script and identify key data processing operations using a knowledge base of python ML APIs~\cite{namaki2020vamsa}.

As a new trend is emerging of automating the pipelines and thus script generation, eg the ``Automated Machine Learning'' or AutoML movement, extracting provenance may become simpler and more precise. Systems in use by commercial providers, such as Databricks' FlowML, are able to capture elements of model generation, to the extent necessary  to make them reproducible.

In the context of Big Data processing, a number of proposals exist to capture fine-grained provenance during data transformation. These include Titian~\cite{Interlandi:2015:TDP:2850583.2850595}, which however is primarily designed to enable debugging of Big Data processing pipelines, with focus on the Spark distributed data processing engine. A recent ``roadmap'' for the role of provenance in Spark-based scientific workflows is available from Guedes et al~\cite{8638375}. Still in the context of map-reduce based systems, ~\cite{lineageSurvey} presents state of the art and research directions for provenance in the context of ``Hadoop as a Service''. 

Finally, it must be noted that the detailed provenance of a process may expose sensitive elements of the data and its transformations. Privacy preservation for provenance is a well-studied topic but few proposals exist that are sufficiently recent or that have come to fruition. Cheney et al.~\cite{10.1007/978-3-319-16462-5_9} provide a quick survey on provenance sanitisation. Amongst the  prototype tools to create views over provenance data (which is represented as a graph), ProvAbs~\cite{Missier2015}  stands out as it is based on the W3 standard PROV data model for describing provenance.





\subsection{Safety \& Security in the Model-Centric Stages}	
\label{ss:model:safety}
Whereas for the data-centric stages we discussed issues in data confidentiality and integrity that are relatively well-understood, in the model-centric stages we come across issues that are traditionally less deeply explored in computing.  We will look at approaches that are directly related to the models and algorithms of a system. Also, adversarial machine learning attacks and defences are discussed in data--centric stages~(see Section~\ref{ss:data:safety}; therefore, we will not discuss them in this section. The technologies mostly relate to the vulnerability as well as the protection in real-life operation, that is, in the Inference stage of the chain of trust in Figure \ref{fig:pipeline}.  
For example, an attacker may target an intrusion detection system that is actively examining a network to detect malicious activities, this means the system is at the inference stage. Such intrusion detection system relies on a set of rules that are fixed and tested based previously in training and testing stages. The attacker is motivated to explore the possibility of finding a network pattern that would bypass the intrusion detection system's set of rules~(the attack that is known as \emph{evasion attacks}~\cite{securitySurvey} in the literature). If the attacker has full knowledge of the intrusion detection system, then the attempts to discover the intrusive network pattern is done under the \emph{white--box} approach. In contrast, the attacker with no knowledge of the model is determined to rely on \emph{black--box} approach. Considering ML model $h$ trained with parameters $\theta$, $h_{\theta})$, white--box attacker would know part or all of either $h$ and $\theta$ and black--box attacker does not have any prior knowledge about the model or its parameters.

The most common threat in the model-centric stages are spoofing attacks and inversion attacks. {\em Spoofing attacks} refer to inserting made-up data to change the outcomes of the model in a manner advantageous to the attacker. Compared to the data poisoning attack, the attack we consider under spoofing change the data in a semantically meaningful way, namely to purposely change the outcome.  Spoofing attacks come in two incarnations, evasion and impersonation. In case of {\em evasion}, the attacker generates an adversarial set of samples to evade detection by the model. These attacks mainly compromise information--security focused applications such as malware~\cite{poisoning2} and spam~\cite{poisoning1} detection systems. The real--world implementation of these attacks has received considerable attention in the literature \cite{biggio2013evasion,zhang2015adversarial,li2014feature,sharif2016accessorize}. {\em Impersonation attacks} spoof the data with the aim to pretend that the data input is from a genuine source. In a typical impersonation scenario, attackers covertly pretend to be the victim. Then, they inject malicious samples to the model as if it is generated by the victim; resulting into different classification outcome than the original data~\cite{laskov2014practical,papernot2017practical,carlini2017towards}. This way, the attackers can gain the same access in the system as the victim`s~\cite{securitySurvey}. These attacks has been implemented in image recognition~\cite{milicheluding}, malware detection~\cite{maiorca2013looking,xu2016automatically} and intrusion detection system~\cite{kayacik2007automatically}. Furthermore, they have been proved to be effective in deep neural networks~(DNN) since they tend to compact the features and thus, the attacker can potentially more easily modify key features to obtain their purpose~\cite{DBLP:journals/corr/MopuriGB17,moosavi2017universal,nguyen2015deep,biggio2014poisoning,sharif2016accessorize}. 

In {inversion attacks} the attacker first gather information about the model operation, e.g., through its application program interface, and then use the insights gained to attack the model~\cite{shokri2017membership,wu2016methodology}.  Inversion attacks can be implemented in white--box or black--box settings, depending on the level of information that the attacker is able collect~\cite{fredrikson2015model}. Such attacks become even more powerful if they utilize data leakage vulnearbilities \cite{fredrikson2015model,fredrikson2014privacy,tramer2016stealing} or misclassification erros \cite{fredrikson2014privacy}. In an interesting demonstration of this attack, Tram\`er et al.~\cite{tramer2016stealing} specifically targeted cloud based machine learning services to design attacks using the confidence values output by the cloud--based models in Amazon and Google.

Countermeasures against such threats are not straightforward. One of the challenges in model--centric security solutions is the diversity of models in machine learning and proposed solutions are usually only applicable to a single model class. However, also the mere application of known techniques such as cryptography to model operations instead of data is not trivial by any stretch. In model-centric solutions calculating over encrypted values relies on {\em homomorphic encryption} \cite{hu2017generating}. In homomorphic encryption, one can perform mathematical operations on encrypted data without the need to decrypt them. The model operates based on a joint effort by multiple parties which have partial knowledge about the system. This way, the whole system operates correctly without leaking sensitive information about its internal functionality and the data to any parties, including the malicious ones. Shokri et al.~\cite{secureCompuation} propose a deep--learning privacy persevering based on homomorphic encryption. They propose a system that works on a server that is honest--but--curious. 
A variety of homomorphic encryption applications, such as multi--party computation~\cite{wang2014compressive,damgaard2012multiparty}, classification over homomorphic encryption data~\cite{aslett2015encrypted}, homomorphic encryption based neural networks~\cite{gilad2016cryptonets} and a homomorphic encryption version of k--means clustering~\cite{yucheng2017investigation}. Homomorphic encryption has large potential in cloud--based machine-learning-as-a-service operations, but efficiency remains a bottleneck. In this context, Ohrimenko et al.~\cite{secureCompuation} is of interest.  It proposes an alterative to homomorphic encryption using a secure processor architecture based on SGX--processors in order to perform various types of machine learning models including neural networks, SVM, matrix factorization, decision--trees and k--means clustering. Their evaluations show that their solution bears relatively low overhead compared to solutions based on homomorphic encryption.

In addition to above confidentiality issues, the literature also proposes a number of solutions to maintain the integrity of the model outcomes. Demontis et al.~\cite{demontis2017yes} present a secure SVM training procedure they call sec--SVM, which learns more evenly--distributed features in linear classifiers. Biggio et al.~\cite{trainingAttack3,trainingAttack4} have a series of solutions of protective algorithms to safeguard the training stage against data poisoning attacks. They use techniques such as random subspace method, bootstrap aggregating and bagging ensembles to search for outliers during model training. In the testing and inference stages most of the solutions are based on game theory, to model the potential behaviours of attacker and defender \cite{securitySurvey}. Globerson et al.~\cite{testingAttack1} use a min--max approach to analyse the impact of all possible feature manipulations in the testing stage for SVM algorithm, further improved by Teo et al.~\cite{teo2008convex} to consider additional attack scenarios. Br\.uckner et al.~\cite{testingAttack3} suggest a novel adversarial prediction system called Stackelberg Games and provide an algorithm based on Nash equilibrium~\cite{bruckner2012static} called NashSVM. Bul\`o et al.~\cite{bulo2016randomized} improve~\cite{testingAttack3} with a randomized strategy selection in their prediction system, thus improving the trade--off options between attack detection and false alarm. 

A quite different approach to preserve the integrity of the model operation is to purposely account for \emph{adversarial} settings in the training, testing and inferring stages. In these approaches one retrains the model with adversarial samples~\cite{securitySurvey,szegedy2013intriguing} so that the retrained model detects anomalies \cite{testingAttack4,testingAttack5}. Goodfellow et al.~\cite{testingAttack6} showed that this can be done reasonably efficiently, reduce the false recognition rate. Of course, the inherent limitation of such approach relies on good adversarial samples, in all permutations, and the methods is therefore not suited for unknown or xero-day attack scenarios. 

A final opportunity to protect against manipulated results is to correct the output of the model, or at least warn for suspicious outputs. Gu et al.~\cite{gu2014towards} construct a deep contractive network that adds a penalty score in the model output, and approach mostly aimed at deep learning models. Papernot et al.~\cite{papernot2016distillation} propose a variant called defensive distillation to defend DNN models, but some concerns about the effectiveness of this method remain ~\cite{carlini2016defensive} arguing about the security of ~\cite{papernot2016distillation}. 


\section{Open Problems}
\label{s:open}
There is a substantial amount of research being invested in each of the four FEAS aspects of building trustworthy machine learning systems: fairness, ethics, auditability and safetry/security/privacy.  For each of the FEAS properties research challenges are being pursued by researchers world-wide, and in this paper we want to highlight the open problems that exist in combining the resulting research contributions to assist data and software engineers in building trustworthy AI-based systems that fulfil the objectives of the policy frameworks discussed in Section \ref{s:policy}. 

\paragraph{Definitions and Implementation of Fairness.} It is probably only natural that, in a fast emerging research area such as trustworthy machine, learning terminology as well as key approaches have not yet converged.  This is most apparent with respect to the fairness definitions we surveyed in Section \ref{ss:fairness}, of which there are tens available \cite{fairnessDefintions}.  It is theoretically impossible to satisfy all fairness definitions at the same time \cite{fairnessImpossibility}. This ambiguity in definitions can have serious implications. A prime example is the well-known COMPAS case, which concerns an approach to prisoner recidivism prediction used in several US states. The reported unfairness across ethnic groups was rebutted by the software company based on its own, different, fairness definition \cite{larson2016we,dressel2018accuracy,feller2016computer}. In addition, reliable practical usage of fairness definitions is not without challenges. Incorrect application of fairness metrics may result in increased discrimination \cite{counterfactual} and fairness metrics may be prone to result instability, that is, a small change in training set may have noticeable impact on the discrimination performance \cite{fairnessImpossibility}. It is clear developers of machine learning systems can easily be left confused about what definitions and associated algorithms and software to apply.  There is a clear open problem in establishing an evidence-based implementation framework for fairness in machine learning, covering definitions of metrics, their semantic subtleties with respect to the discrimination they address, and the associated real-life implementation pitfalls for builders of machine learning systems. 

\paragraph{Integration of Trustworthiness and Accuracy.} Machine learning algorithms are typically evaluated with respect to accuracy and performance first, which is understandable if models are developed isolated form the context of a real-life system. When using machine learning systems in real-life, trustworthiness often becomes as or more important than accuracy and performance.  If the FEAS properties are considered too late in the design of algorithm and system, it may not be possible to adjust the machine learning system satisfactory. There is no general mechanism to balance accuracy and fairness for a set of ML models, so far the research on this topic is limited to specific models and conditions. This implies a clear research challenge in trustworthy machine learning, namely the integrated consideration of accuracy, performance and FEAS properties. This requires development of trade-off approaches with respect to metrics representing the different objectives, and the development of combined metrics. An associated open problem is the application of known trustworthiness techniques to AI-based systems in general and machine-learning pipelines in particular. In particular, systematic treatments such as advocated by \cite{Avizienis04} for designing dependable systems needs to be further expanded for AI-based systems.   

\paragraph{Data Engineering Pathways to Implement Policy Frameworks.} The chain of trust introduced in Section \ref{s:chain} asks a data engineer and system builder to consider a sequence of data processing steps, which requires to transform the datasets in ones stage into a form that can be used as input to a later stage. There currently are no well-established data and software engineering approaches for such data transformation in relation to achieving the desired combined accuracy, performance and FEAS properties. Not only are there gaps in knowledge of how individual data or processing choices can affect the outcome, but the formal analysis of how to combine data and processing has not been much explored. There is a need for concerted efforts in creating tools that help service providers and machine learning system builders communicate and to help the latter engineer such requirements into model design. This requires research in formalisation and tooling, which would beneficially impact all stakeholders, from data scientists to service providers and end users.

\paragraph{Integration of Social Science Theories and Perspectives.} In this paper, we discussed, on the one hand, the various policy frameworks that are emerging and, on the other hand, the relevant FEAS technologies that are available. To understand best how to bridge societal concerns and technological possibilities research should be informed by multi-disciplinary perspectives. As an example, \cite{toreini2020relationship} conducts initial research in this space in relation to trust in AI-based services. In that work, social science theories for trust development are linked up with technology quality requirements. Evidence-based research in social-science, psychology and human-computer interaction has the potential to steer technology development and demonstrate its abilities or limitations in achieving societal goals. Through quantitative as well as qualitative research the impact of technologies with respect to their societal impact can be measured and better understood.    

\paragraph{Education.}  Veale et al.~\cite{veale2018fairness} interview 27 public sector machine learning practitioners in five countries about their understanding of public values such as fairness and accountability. Their results confirm the disconnect between individual perspectives and technological properties and point out the challenges that come with this in developing ethical machine learning systems. Similarly, Holstein et al.~\cite{holstein2019improving} interview 25 machine learning practitioners and survey 267 more to systematically investigate the implementation of fair machine learning by commercial product teams. They also highlight the difference between real--world challenges and the provided solutions.  Improving this situation could start with education of all stakeholders in machine learning systems, including software and data engineers, for instance through curriculum development such as demonstrated by \cite{bates2020integrating}.  


\section{Conclusion}
\label{s:conclusion}
This paper reviews the landscape and the most pertinent technologies required for building machine learning systems that aim to fulfil the objectives of the many AI-related policy frameworks that are emerging in relation to fairness and ethics.  The paper identifies four main system properties to implement such systems, namely fairness, explainability, auditability and safety/security (FEAS), and provides a review of the main trends and technology categories in each of these.  It surveys the FEAS technologies with respect to all stages of the machine learning chain of trust, subdivided in data-centric and model-centric technologies, respectively.  The paper argues that in future work, the emphasis should not only be on individual FEAS technologies but particularly on how these technologies integrate with accuracy and efficiency consideration, can be translated in data and software engineering tools that integrate concerns across the various stages of the machine learning life cycle and in cross-disciplinary research to more profoundly connect FEAS technologies with the AI policy framework objectives they ultimately set out to achieve. 

\begin{acks}
Research supported by UK EPSRC, under grant EP/R033595, "Trust Engineering for the Financial Industry", \url{https://gow.epsrc.ukri.org/NGBOViewGrant.aspx?GrantRef=EP/R033595/1}.
\end{acks}

\bibliographystyle{ACM-Reference-Format}
\bibliography{technical.bib,social.bib}



\end{document}